\documentclass[letterpaper, 10 pt, journal, twoside]{IEEEtran}
\usepackage{amsmath,amsfonts}
\usepackage{algorithmic}
\usepackage{array}
\usepackage[caption=false,font=normalsize,labelfont=sf,textfont=sf]{subfig}
\usepackage{textcomp}
\usepackage{stfloats}
\usepackage{url}
\usepackage{verbatim}
\usepackage{graphicx}
\usepackage{booktabs}
\usepackage{multirow} 
\usepackage[colorlinks=true, linkcolor=blue, citecolor=blue, urlcolor=blue]{hyperref}
\def\BibTeX{{\rm B\kern-.05em{\sc i\kern-.025em b}\kern-.08em
    T\kern-.1667em\lower.7ex\hbox{E}\kern-.125emX}}
\usepackage{balance}
\begin{document}

\title
{	
{OpenMulti: Open-Vocabulary Instance-Level Multi-Agent Distributed Implicit Mapping}
    
\thanks{This work is supported by the National Natural Science Foundation of China under Grant 92370203, 62473050, 62233002, Beijing Natural Science Foundation Undergraduate Research Program QY24180. \textit{(Corresponding authors: Yufeng Yue.)}
}
}

\author{
{Jianyu Dou$^{1}$, Yinan Deng$^{1}$, Jiahui Wang$^{1}$, Xingsi Tang$^{1}$, Yi Yang$^{1}$ and Yufeng Yue$^{1*}$}

\thanks{All authors are with School of Automation, Beijing Institute of Technology, Beijing, 100081, China.}
}




\maketitle



\begin{abstract}

Multi-agent distributed collaborative mapping provides comprehensive and efficient representations for robots. However, existing approaches lack instance-level awareness and semantic understanding of environments, limiting their effectiveness for downstream applications.
To address this issue, we propose OpenMulti, an open-vocabulary instance-level multi-agent distributed implicit mapping framework.
Specifically, we introduce a Cross-Agent Instance Alignment module, which constructs an Instance Collaborative Graph to ensure consistent instance understanding across agents.
To alleviate the degradation of mapping accuracy due to the blind-zone optimization trap, we leverage Cross Rendering Supervision to enhance distributed learning of the scene. Experimental results show that OpenMulti outperforms related algorithms in both fine-grained geometric accuracy and zero-shot semantic accuracy. In addition, OpenMulti supports instance-level retrieval tasks, delivering semantic annotations for downstream applications.
The project website of OpenMulti is publicly available at \url{https://openmulti666.github.io/}.

\end{abstract}


\section{INTRODUCTION}

\IEEEPARstart{S}{cene} reconstruction is a fundamental capability for agents to execute downstream tasks effectively \cite{nanshen}. For a single agent, achieving a comprehensive environmental understanding necessitates full-area scanning, a process that becomes particularly challenging in complex environments. To address this, multi-agent collaborative mapping has emerged as a critical solution. Existing approaches \cite{hdccsom,tangtang} often rely on classical explicit representation techniques, where agents exchange voxel or point cloud data generated individually. However, since these methods typically consist of discrete map units, it is challenging to achieve an optimal balance between mapping accuracy and transmission costs \cite{macim}.

\begin{figure}[t]
    \centering
    \includegraphics[width=0.45\textwidth]{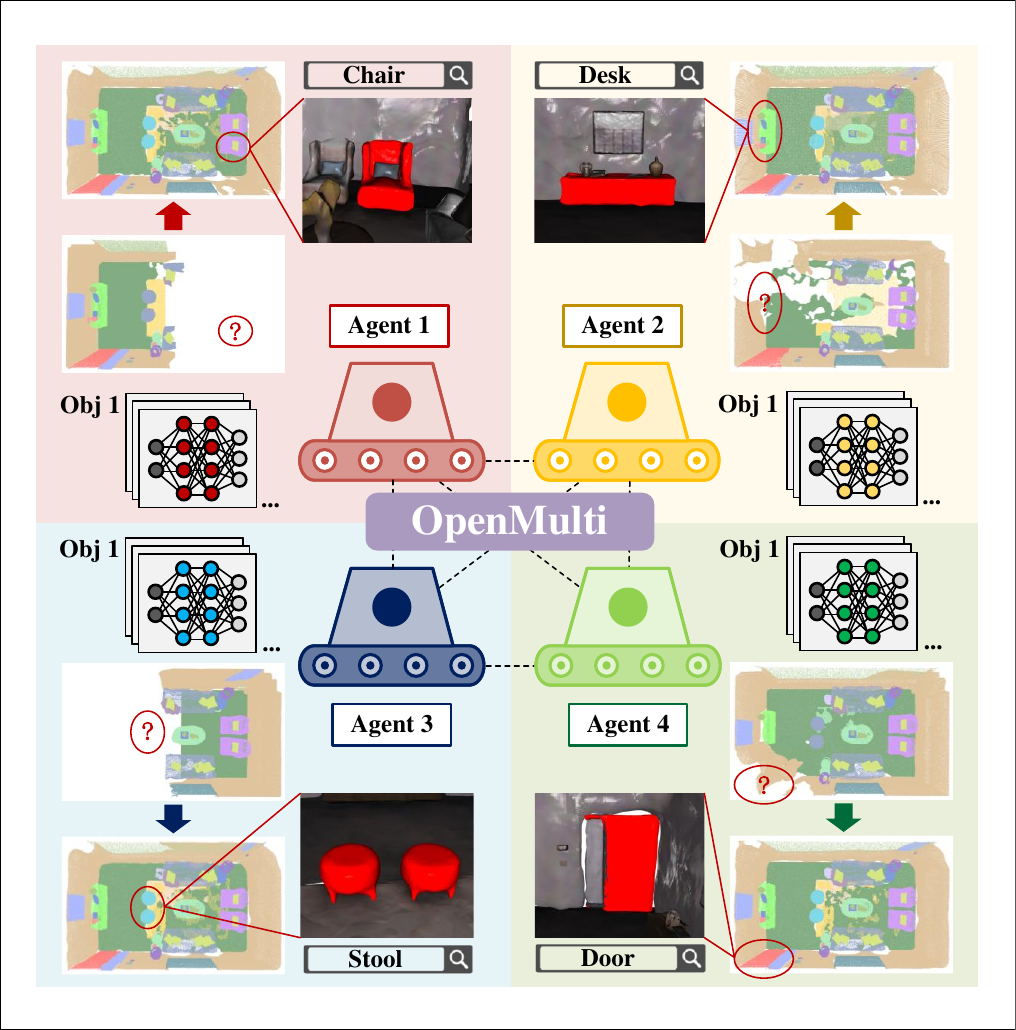}
    
    \caption{We propose OpenMulti, a framework for constructing consistent open-vocabulary implicit maps across multiple agents. OpenMulti facilitates information exchange between agents, effectively compensating for regions not observed by individual agents.
    Additionally, OpenMulti supports downstream tasks, including open-vocabulary instance retrieval.}
    
    \label{fig:1}
    \vspace{-5pt}
\end{figure}

To address these challenges, recent approaches leverage implicit mapping techniques \cite{nerf} to enable lightweight multi-agent mapping. These methods \cite{dinno,macim,dinerf,ramen} achieve map fusion by transmitting implicit network parameters, effectively decreasing communication overhead. However, they typically do not possess semantic comprehension capability and instance-level awareness, restricting their utility in downstream task-intensive applications, such as object navigation or semantic retrieval. Consequently, there is a pressing demand for open-vocabulary instance-level multi-agent implicit mapping to facilitate a more comprehensive and nuanced environmental understanding. However, achieving this goal involves addressing two key challenges.

The first challenge is \textbf{how to achieve instance alignment across agents to enhance consistent scene understanding.}
Due to varying viewpoints, multiple agents may perceive instances in the same scene with different attributes. These discrepancies can be categorized into four main types: semantic errors, over-segmentation, viewpoint loss, and under-segmentation. For example, when two cups are placed side by side, their similar appearance and potential occlusion may lead to their misidentification as a single instance from certain agent viewpoints.
To address these front-end perception ambiguities, it is essential to incorporate segmentation information from other agents. OpenMulti introduces Cross-Agent Instance Alignment module that constructs an Instance Connectivity Graph. By leveraging spatial overlaps between instances and utilizing confidence probabilities to select reference agents, the module refines and updates instance information across agents. This approach enables OpenMulti to achieve instance alignment and unification across agents, laying the groundwork for subsequent instance-level implicit mapping.

The second challenge is \textbf{how to enhance the accuracy of multi-agent collaborative mapping under the constraint of low communication cost.}
To decrease communication overhead, recent approaches \cite{dinno,macim,dinerf} transmit only implicit network parameters for mutual learning, enabling lightweight communication. However, this results in the blind-zone optimization trap, where agents lack supervision in occluded regions and parameter-level consistency alone fails to guarantee coherent rendering, often causing artifacts like geometric holes and discontinuities. 
OpenMulti addresses this issue by proposing Cross Rendering Supervision, which enhances scene reconstruction by calculating depth differences from the same ray direction. Notably, OpenMulti does not directly transmit raw depth data. Instead, it shares ray direction alongside implicit parameters and leverages network rendering to obtain depth information from other agents' ray directions. This shift from abstract parameter optimization to intuitive data supervision significantly improves mapping accuracy.

In summary, OpenMulti introduces a novel framework for instance-level open-vocabulary implicit representation in multi-agent collaborative mapping. As illustrated in Fig. \ref{fig:1}, OpenMulti achieves instance alignment across multiple agents while ensuring consistent global semantic understanding. By enabling retrieval tasks through feature similarity calculations, the instances are highlighted within the scene, delivering semantic annotations for potential downstream applications.
Our contributions are summarized as follows:
\begin{itemize}
    \item We propose OpenMulti, an open-vocabulary, instance-level, multi-agent distributed implicit mapping framework that constructs consistent and collaborative maps across multiple agents.
    \item We introduce the Cross-Agent Instance Alignment module, a confidence-based semantic unification strategy that ensures consistent instance recognition among agents.
    \item We propose Cross Rendering Supervision to address the blind-zone optimization trap, thereby further enhancing the accuracy of multi-agent collaborative mapping.
    \item Extensive experiments across diverse scenes demonstrate that OpenMulti achieves superior geometric and semantic accuracy while supporting downstream tasks such as instance retrieval.
\end{itemize}

\section{RELATED WORKS}

\subsection{Open-Vocabulary Mapping}

Recent advances in Visual Language Models (VLMs) have enabled mapping systems to perform open-vocabulary semantic understanding from multimodal data \cite{OpenVox}. Early approaches utilized depth measurements to directly project 2D point-wise VLM features into 3D space\cite{vlmaps},\cite{conceptfusion},\cite{OpenScene}. However, such methods lack instance-level awareness, significantly limiting their effectiveness in interactive downstream applications.

Subsequent research has introduced instance-level semantic maps to achieve more refined scene understanding. For instance, ConceptGraph \cite{conceptgraphs} and OpenGraph \cite{opengraph} integrate object point clouds frame-by-frame by leveraging geometric and feature similarity. While these methods enable agents to extract instance-level information from diverse map regions, they predominantly rely on point clouds, voxels, or other discrete representations as map structures, resulting in high storage costs and insufficient map continuity.

In recent years, NeRF \cite{nerf} has emerged as a research hotspot due to its photorealistic rendering quality and lightweight implicit representation. Several studies have explored combining NeRF with open-vocabulary semantic capabilities. For instance, LERF \cite{lerf} pioneered the integration of multi-scale CLIP features into NeRF, albeit at the cost of reduced training and inference efficiency. Similarly, 3D-OVS \cite{3dovs} introduces a correlation distribution alignment loss to enhance segmentation and optimize semantic feature domains. OpenObj \cite{openobj} utilizes the over-segmentation capability of the SAM model to extract pixel-aligned part-level features, enabling finer-grained object understanding in scenes.

Despite these advancements, existing open-vocabulary mapping methods remain confined to single-agent systems, suffering from low efficiency and incomplete scene understanding and modeling due to perceptual limitations.

\begin{figure*}[t]
    \centering
    \includegraphics[width=1\textwidth]{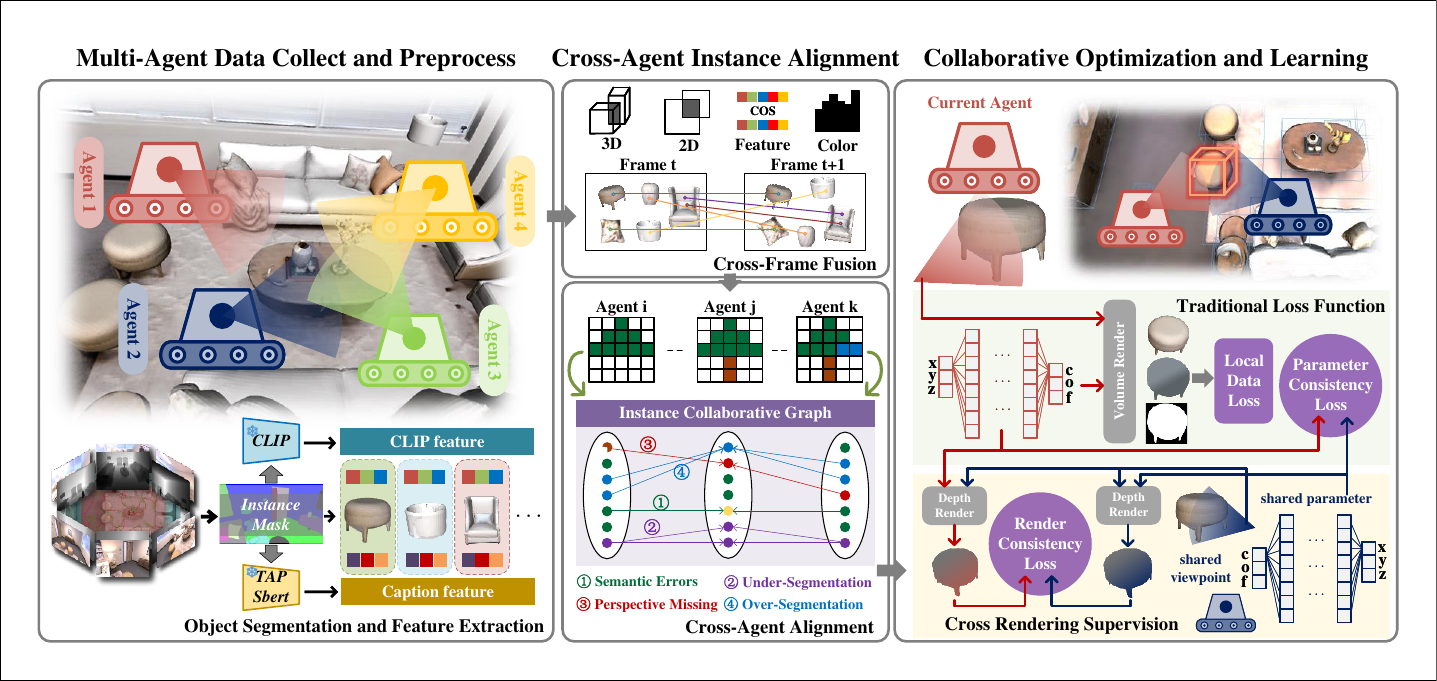}
    \caption{The framework of OpenMulti consists of three main components: Multi-Agent Data Collect and Preprocess, Cross-Agent Instance Alignment, and Collaborative Optimization and Learning.}
    \label{fig:2}
    \vspace{-5pt}
\end{figure*}


\subsection{Multi-agent Collaborative Mapping}

To enhance scene reconstruction efficiency and leverage the benefits of cross-perspective cognition, research on multi-agent collaborative mapping is essential. Early multi-agent collaborative mapping methods relied on sharing local point cloud maps \cite{tangtang} or voxel maps \cite{hdccsom} among agents. However, these approaches impose significant communication overhead, making real-time transmission challenging.


Alternative approaches model collaborative implicit mapping as distributed optimization problems, reducing communication overhead by sharing key data and ensuring consistent mapping across agents. DiNNO \cite{dinno} represents a pioneering effort in this direction, laying the groundwork for subsequent research. MACIM \cite{macim} extends the DiNNO method to ESDF mapping tasks, incorporating consistency loss to achieve cross-platform parameter unification. Similarly, Di-NeRF \cite{dinerf} not only unifies parameters but also optimizes the relative poses between agents, enabling global mapping and localization. RAMEN \cite{ramen} derives an uncertainty-weighted decentralized C-ADMM algorithm, enabling more robust and accurate multi-agent implicit mapping.

Concurrent works like MAGiC-SLAM \cite{magicslam} and HAMMER \cite{hammer} investigate multi-agent centralized mapping systems by employing 3D Gaussian Splatting \cite{3dgs} as the map structure. However, these methods lack instance-level awareness, significantly diminishing their practicality in real-world robotic tasks. As their code is not publicly available, they are excluded from the baselines in experiments.

\section{OpenMulti}

\subsection{Framework Overview} 

Fig. \ref{fig:2} illustrates the OpenMulti framework, where multiple agents are deployed in a common scene. First, each agent performs data collection and processing, capturing RGB and depth images, recording poses, and performing instance segmentation along with feature extraction. The Cross-Agent Instance Alignment module then constructs an Instance Connectivity Graph to ensure semantic consistency. 
Finally, the instance-level NeRF network maintained by each agent is optimized using a traditional loss function and the cross-rendering supervision module, which addresses the blind-zone optimization trap and enhances the accuracy of collaborative multi-agent implicit mapping.

\subsection{Multi-Agent Data Collect and Preprocess}

In our setup, each agent independently collects environmental data, which remains private throughout the mapping process. This includes RGB images $I_k^c=\{{I}^c_{k,t}\}$ , depth images $I_k^d=\{{I}^d_{k,t}\}$, and camera poses $\Omega_k=\{{\Omega}_{k,t}\}$ for k-th agent $a_k$ at all timestamps $\{t\}$.
This module follows the fronted processing approach of OpenObj \cite{openobj}, using the advanced instance segmentation tool CropFormer \cite{entity} model to obtain instance masks $\{m_{k,t,m}^{inst}\}$. We then extract semantic features using zero-shot foundation models: CLIP model encodes mask-cropped images into CLIP features $\{f_{k,t,m}^{clip}\}$ representing appearance information, while TAP model generates mask descriptions that SBERT encodes into caption features $\{f_{k,t,m}^{caption}\}$ capturing textual semantics. These features offer complementary perspectives, with one focusing on visual appearance understanding and the other on semantic reasoning about the environment. Here we employ a Code Book approach to store both types of semantic features, where each instance ID $m$ uniquely indexes a set of semantic features.


\subsection{Cross-Agent Instance Alignment} \label{Section.C}

This module comprises three components: \textbf{Cross-Frame Fusion}, \textbf{Confidence-Based Instance Propagation}, and \textbf{Cross-Agent Alignment}.

\textbf{Cross-Frame Fusion:}  
In this stage, each agent processes all the masks $m_{k,t,m}^{inst}$ it collects, ensuring semantic fusion over time. Following the approach of OpenObj \cite{openobj}, Cross-Frame Fusion is performed in two phases: the Coarse Clustering Phase and the Fine Clustering Phase. This results in the unified instance $O_{k,i}$ and associated semantic feature $f_{k,i}^{clip},f_{k,i}^{caption}$ for agent $a_k$.

\textbf{Confidence-Based Instance Propagation:} A core challenge in multi-agent mapping lies in handling inconsistent segmentation granularity and viewpoint-dependent errors. To address this, we introduce a confidence estimation mechanism that quantifies the reliability of each instance based on its spatial location within the image. Instances appearing near image borders are prone to occlusions and segmentation inaccuracies, resulting in lower confidence.

For each mask ${O}_{k,t,i}$ in image $I_{k,t}$, we compute the per-frame confidence probability:
\begin{equation}
{{C}_{k,t,i}}=\left\{ \begin{matrix}
   0 & {{O}_{k,t,i}}\notin I_{k,t}^{c}  \\
   \frac{S({{O}_{k,t,i}})}{S(I_{k,t}^{c})} & {{O}_{k,t,i}}\notin I_{k,t}^{c,center}  \\
   1 & {{O}_{k,t,i}}\in I_{k,t}^{c,center}
\end{matrix} \right.
\end{equation}
where $I_{k,t}^{c,center}$ indicates the central region of the image. $S(\cdot)$ denotes the number of pixels within a mask or the whole image.  
The final confidence for an instance is averaged over time:
\begin{equation}
C_{k,i}=\frac{1}{T}\sum_{t=1}^{T}C_{k,t,i}
\end{equation}

\textbf{Cross-Agent Alignment:} In this phase, agents exchange information to construct the Instance Collaborative Graph to execute Cross-Agent Alignment.

\begin{figure}[t]
    \centering
    \includegraphics[width=0.48\textwidth]{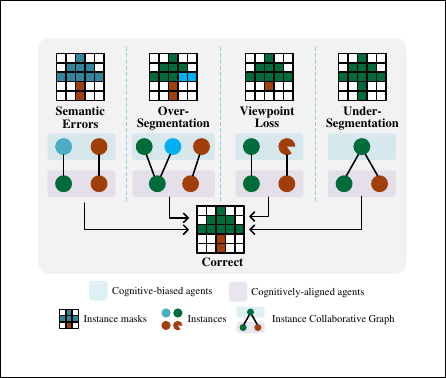}
    \caption{Key issues addressed by Cross-Agent Alignment. The upper and lower matrices represent erroneous and correct instance masks respectively, which are connected through the intermediate Instance Collaborative Graph to ultimately rectify incorrect instance interpretations.}
    \label{fig:3}
    \vspace{-5pt}
\end{figure}

Fig. \ref{fig:3} illustrates several issues encountered in cross-frame fusion. To address them, OpenMulti introduces the Instance Collaborative Graph, where each instance is represented as a vertex, and edges are constructed to indicate instances belonging to the same instance (as shown in Fig. \ref{fig:3}).
The construction process is as follows: First, all instances $O_{k,i}$ from different agents are converted into point clouds $P_{k,i}$ using depth images $I^{d}_{k}$ and camera poses $\Omega^{d}_{k}$. As high-precision calculations are not required, we downsample the point clouds to reduce communication overhead between agents. Next, the spatial overlap ratios IoU and IoB are computed between instances from different agents to assess their spatial consistency in the 3D scene. We compute them by evaluating nearest neighbor distances. For each point in $P_{k_1,i}$, if the nearest point in $P_{k_2,j}$ lies within a preset distance threshold (0.01m in our experiment), it is considered overlapping. The formulas are:

\begin{equation}
    IoU_{k_1,i}^{k_2,j}=\frac{|P_{k_1,i}\cap P_{k_2,j}|}{|P_{k_1,i}\cup P_{k_2,j}|}
\end{equation}
\begin{equation}
    IoB_{k_1,i}^{k_2,j}=\frac{|P_{k_1,i}\cap P_{k_2,j}|}{|P_{k_1,i}|}
\end{equation}

Based on the spatial overlap ratios, thresholds are applied to determine whether two instances $O_{k1,i}$ and $O_{k2,j}$ belong to the same instance. If they do, an edge is created between them. 
This process is repeated for all instance pairs, resulting in the construction of the Instance Collaborative Graph.

Starting from a selected instance within an agent, we iteratively traverse connected edges to identify all related instances, forming a cluster. Within each cluster, the instance with the highest confidence is chosen as the reference agent to update others' information, thereby rectifying inconsistent semantic outcomes.

1. For Semantic Errors and Viewpoint Loss, the semantic features and instance IDs of lower-confidence instances are directly replaced by those of the reference instance, while keeping their original masks unchanged.

2. For Over-Segmentation, clusters are formed by grouping overlapping masks with similar semantic features. All masks within a cluster are merged into a unified mask, and the semantic features and instance IDs of the reference instance are propagated to the merged result.

3. For Under-Segmentation, the reference agent’s point cloud data is back-projected onto the 2D image space of the current agent to generate refined segmentation masks. These newly segmented masks are then individually aligned to the reference instance in terms of both features and instance IDs, ensuring consistent cross-agent semantic interpretation.

The updated mask information is denoted as $\overline{m}_{k,t,m}^{inst}$. After aforementioned semantic correction process, the Graph achieves an injective mapping state, meaning that each instance within one agent is uniquely and consistently mapped to a corresponding instance in the other agents. This establishes a one-to-one alignment across all agents, completing the Cross-Agent Alignment process.

\subsection{Collaborative Optimization and Learning}

In OpenMulti, some instance-level NeRF networks are constructed rather than the global network as in \cite{macim}. To optimize network parameters, OpenMulti calculates two traditional loss terms for distributed collaborative scene mapping. Furthermore, a Cross Rendering Supervision module is introduced to further enhance mapping accuracy through Render Consistency Loss.

\textbf{Instance-level NeRF:} OpenMulti constructs a NeRF network $F_{k,i}(\vec{x})$ for each instance in all agents. Unlike traditional NeRF modules, OpenMulti builds a small MLP network for each instance rather than a global network for the entire scene. This approach allows for individual optimization of each instance and enables effective information sharing among agents $\{a_k\}$, achieving higher mapping accuracy. The MLP network for the i-th instance in agent $a_k$ can be written as:

\begin{equation}
    F_{k,i}(\vec{x})\rightarrow (\vec{c},\sigma)
\end{equation}
where $\vec{x}=(x,y,z)$ represents the coordinates within instance $O_{k,i}$, $\vec{c}=(r,g,b)$ represents the color, $\sigma$ represents the volume density. Since we have already extracted complete semantic features for each instance in the preceding section and no longer require further feature refinement within instances, we continue using the Code Book for feature storage instead of designing semantic features as network outputs.

Based on the camera pose $\Omega$ and viewing direction $\vec{d}$, we sample rays $\vec{r}$ to obtain $N$ points $\{\vec{x_l}\}_{l=1}^N$. The probability $T_p$ of the ray terminating at point $\vec{x_p}$ is defined as:

\begin{equation}
    T_p=\sigma (\vec{x_p})\prod\limits_{q<p} (1-\sigma (\vec{x_q}))
\end{equation}

Based on the termination probability $T_p$, the following rendered occupancy, depth and color are obtained:

\begin{equation}
    \hat{O}(\vec{r})=\displaystyle \sum_{p=1}^N T_p\quad \hat{D}(\vec{r})=\displaystyle \sum_{i=p}^N T_p d_p\quad \hat{C}(\vec{r})=\displaystyle \sum_{i=p}^N T_p c_p
\end{equation}
where $d_p$ is computed as the distance from $x_p$ to the camera pose, and $c_p$ is obtained by directly querying the NeRF network.

For final rendering, we first identify the instance intersected by each pixel's cast ray and perform point sampling along the ray within each instance. By querying the instance-level NeRF, we obtain the color and density values at sampled points. Finally, we employ differentiable volume rendering to composite multi-instance contributions through depth-aware ranked integration, outputting the final synthesized color for each pixel \cite{vmap}.

\textbf{Traditional Loss Function:} OpenMulti calculates two traditional loss items to optimize the NeRF network in a distributed manner.

1) Local Data Loss: Each agent independently computes the discrepancy between the rendered result and its raw data to derive the Local Data Loss. By sampling the corresponding pixel $\vec{p}=[u,v]$ on the 2D image based on the rendering ray $\vec{r}$, each loss value for the k-th agent $a_k$ is calculated as follows:

\begin{equation}
    L_k^{occ} = \sum\limits_{\vec r \in {a_k}} {|{{\hat O}_k}(\vec r) - \overline{m}_k^{inst}(\vec p)|}
\end{equation}
\begin{equation}
    L_k^{depth} = \overline{m}_k^{inst}(\vec p) \cdot \sum\limits_{\vec r \in {a_k}} {|{{\hat D}_k}(\vec r) - I_k^d(\vec p)|}
\end{equation}
\begin{equation}
    L_k^{color} = \overline{m}_k^{inst}(\vec p) \cdot \sum\limits_{\vec r \in {a_k}} {|{{\hat C}_k}(\vec r) - I_k^c(\vec p)|} 
\end{equation}

The Local Data Loss for the k-th agent $a_k$ is obtained by summing the weighted losses as follows:

\begin{equation}
    L_k^{data} = {\lambda _1}L_k^{occ} + {\lambda _2}L_k^{depth} + {\lambda _3}L_k^{color}
\end{equation}

\begin{figure}[t]
    \centering
    \includegraphics[width=0.5\textwidth]{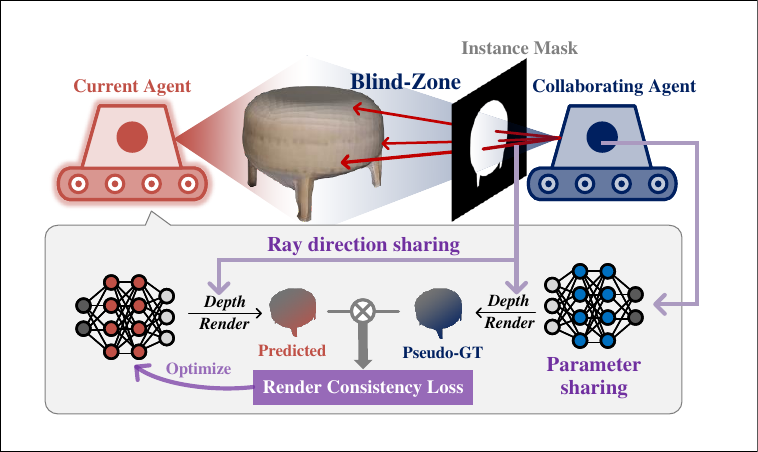}
    \caption{The collaborating agent shares its ray direction with the current agent, enabling both to perform depth rendering under these shared rays and compute the Render Consistency Loss, thereby optimizing the current agent's NeRF network.}
    \label{fig:render}
    \vspace{-5pt}
\end{figure}

2) Parameter Consistency Loss: Owing to the limited scene exploration by a single agent $a_k$, comprehensive environmental understanding is often unattainable. Thus, leveraging information from other agents for distributed optimization is essential. To enhance transmission efficiency, OpenMulti exclusively exchanges NeRF network parameters among agents and formulates a Parameter Consistency Loss to update the parameters:

\begin{equation}
    L_k^{con} = {\sum\limits_{\bar k \in rid(k)} {||{\theta ^k} - {\theta ^{\bar k}}||} ^2}
\end{equation}
where $rid(k)$ denotes the set of all agents excluding $a_k$.

This approach was first proposed in DiNNO \cite{dinno}, then improved and applied to the ESDF mapping in MACIM \cite{macim}, and is now applied to instance-level NeRF mapping in this work, yielding similarly strong results.

\textbf{Cross Rendering Supervision:} As shown in Fig. \ref{fig:render}, OpenMulti introduces Cross Rendering Supervision module to mitigate the blind-zone optimization trap arising from Traditional Loss Function. This module enhances the distributed learning of the scene, thereby further improving mapping accuracy.

1) Blind-Zone Optimization Trap: For the blind-zone of a given instance, an agent lacks visual observations and therefore cannot receive image-based supervision, ultimately resulting in meaningless outputs in those regions. To mitigate this, traditional loss functions employ parameter consistency losses by aligning network parameters across agents. However, due to the highly non-convex and multi-modal nature of NeRF optimization, enforcing parameter-level similarity alone does not guarantee consistent rendering outputs. Especially in large-scale or textureless areas, this can easily lead the optimization to converge to sub-optimal basins, manifesting as geometric holes and discontinuities in the reconstructed map — a phenomenon we refer to as the Blind-Zone Optimization Trap.


2) Render Consistency Loss:  To mitigate the Blind-Zone Optimization Trap, OpenMulti introduces a Render Consistency Loss, which enforces rendering supervision through shared ray directions. As shown in Fig. \ref{fig:render}, collaborating agent shares the current frame's ray directions (by sharing poses and performing consistent ray sampling in blind zones), serving as virtual ray directions for the current agent. Both agents simultaneously perform point sampling along the virtual ray and conduct depth rendering with the NeRF network. The discrepancy between the two depth values is computed to derive the final Render Consistency Loss. The specific formulation is as follows:

\begin{equation}
    L_k^{rend} = \sum\limits_{\bar k \in rid(k),\vec r \in {a_{\bar k}}} {|{{\hat D}_k}(\vec r) - {{\hat D}_{\bar k}}(\vec r)|}
\end{equation}

This process provides direct supervision on rendering outputs in blind zones, guiding the optimization toward geometrically consistent reconstructions and effectively mitigating the Blind-Zone Optimization Trap, thereby improving the mapping accuracy of our method.

\section{Experiment}
\subsection{Experimental Setup}

\textit{Implementation Details:} OpenMulti and all multi-agent baselines simulate the collaborative distributed mapping algorithm on a single server equipped with RTX 4090 GPUs. OpenMulti leverages the PyTorch framework to manage the NeRF network and utilizes the ADAM optimizer for network optimization.

\textit{Evaluation Scenes:} In experiments, OpenMulti selects 8 simulated scenes from Replica \cite{replica} (room\_0, room\_1, room\_2, office\_0, office\_1, office\_2, office\_3, office\_4) and 5 real-world scenes from ScanNet \cite{scannet} (scene0030\_02, scene0220\_02, scene0592\_01, scene0673\_04, scene0692\_02) as the evaluation scenes. For each scene, four data sets are collected to simulate four agents' observations, ensuring comprehensive coverage of the scene.

\textit{Comparison Baseline:} For geometric evaluation, we compare OpenMulti with two SOTA collaborative mapping methods \textbf{MACIM} \cite{macim}, Di-NeRF \cite{dinerf}, and two distributed optimization algorithms DiNNO \cite{dinno}, and DSGD \cite{dsgd}. For the optimization algorithms \cite{dinno,dsgd}, we replace our parameter consistency loss with their distributed loss functions, implementing them as variants \textbf{DSGD\_Instance (DS.I.)} and \textbf{DiNNO\_Instance (Di.I.)} in our framework.
In our experiments, we omitted the localization step in Di-NeRF \cite{dinerf} and employed ground truth poses for a fair comparison, designating this variant as \textbf{Di-NeRF* (Di-N.*)}. For semantic evaluation, we compared OpenMulti with single-agent open-vocabulary approaches, including \textbf{LERF (L.)} \cite{lerf}, \textbf{3D-OVS (3D.)} \cite{3dovs}, \textbf{ConceptGraph (C.G.)} \cite{conceptgraphs}, and \textbf{OpenObj (O.O.)} \cite{openobj}. OpenObj serves as the single-agent baseline, which OpenMulti extends, and is denoted as “OpenSingle” for mapping efficiency comparison in Section~\ref{multivssingle}.

\textit{Evaluation Metric:} For geometry, following prior work \cite{vmap}, we use Accuracy, Completion, and Completion Ratio as 3D scene-level reconstruction metrics. For semantics, F-mIOU and F-mACC are employed to assess semantic accuracy against ground truth. For multi-agent algorithms, the final evaluation results are derived by averaging the metrics across all agents.
\begin{figure*}[t]
    \centering
    \includegraphics[width=1\textwidth]{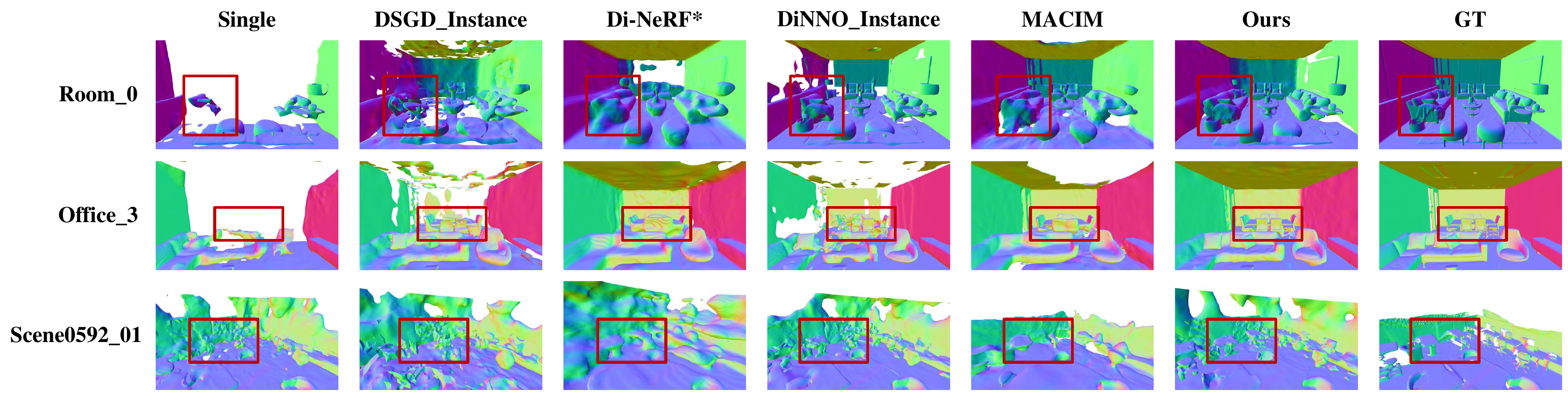}
    \caption{Qualitative comparison of 3D geometric reconstruction. OpenMulti demonstrates enhanced mapping accuracy over other collaborative mapping methods or distributed optimization algorithms. Red boxes indicate areas with notable performance gains.}
    \label{fig:geometric}
    \vspace{-5pt} 
\end{figure*}

\begin{figure*}[t]
    \centering
    \includegraphics[width=1\textwidth]{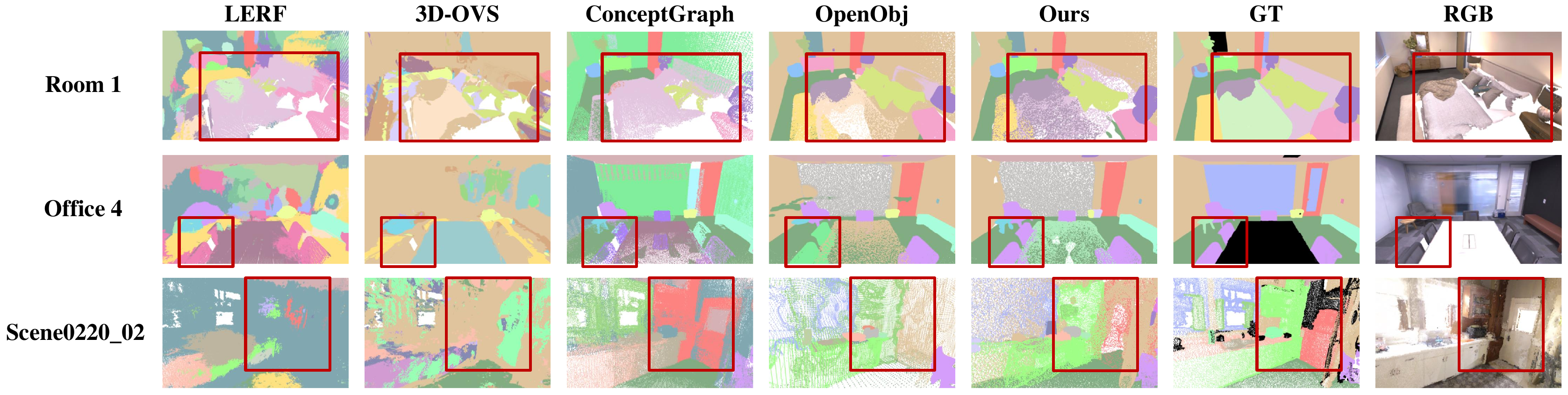}
    \caption{Qualitative comparison of zero-shot 3D semantic segmentation. As a multi-agent distributed semantic mapping method, OpenMulti outperforms centralized open-vocabulary mapping methods in most scenes. Red boxes indicate areas with notable performance gains.}
    \label{fig:semantic}
    \vspace{-10pt} 
\end{figure*}

\subsection{3D Geometric Evaluation}

Fig. \ref{fig:geometric} and Tab. \ref{table:geometric}. present the qualitative and quantitative evaluations of 3D geometric reconstruction. We showcase the results from a random one of the four agents.  
While MACIM and Di-NeRF* employ global map construction, they struggle to capture scene details and may merge adjacent instances. 
DiNNO\_Instance and DSGD\_Instance apply distributed optimization algorithms to instance-level mapping but face ambiguity when multiple agents observe overlapping instances, resulting in reduced mapping accuracy. 
Comparatively, OpenMulti mitigates the blind-zone optimization trap by introducing the Cross Rendering Supervision module in the optimization process, achieving mapping performance superior to other methods.

\begin{table}[!t]
\centering
\scriptsize
\caption{Geometric Reconstruction Results}
\label{table:geometric}
\renewcommand\arraystretch{1.1}
\setlength{\tabcolsep}{1.3mm}
\begin{tabular}{ccccccc}
\toprule
 & \textbf{Single} & \textbf{DS.I.} & \textbf{Di-N.*} & \textbf{Di.I.} & \textbf{MACIM} & \textbf{Ours} \\ \midrule
\textbf{Acc. [cm]$\downarrow$} & 03.32 & 04.41 & 03.46 & 03.66 & 06.93 & \textbf{02.47} \\ 
\textbf{Comp. [cm]$\downarrow$} & 33.01 & 08.04 & 08.48 & 08.04 & 07.10 & \textbf{03.52} \\ 
\textbf{Comp. Ratio [$<$5cm\%] $\uparrow$} & 53.18 & 54.05 & 59.09 & 72.92 & 72.37 & \textbf{81.52} \\ \bottomrule
\end{tabular}
\end{table}

\begin{table}[!t]
\centering
\scriptsize
\caption{Semantic Segmentation Results}
\label{table:semantic}
\renewcommand\arraystretch{1.1}
\setlength{\tabcolsep}{2.3mm}
\begin{tabular}{cccccc}
\toprule
 & \textbf{LERF} & \textbf{3D-OVS} & \textbf{ConceptGraph} & \textbf{OpenObj} & \textbf{Ours} \\ \midrule
\textbf{F-mIoU$\uparrow$} & 18.39 & 03.18 & 28.87 & 44.10 & \textbf{46.18} \\ 
\textbf{F-mAcc$\uparrow$} & 24.04 & 03.71 & 35.67 & 49.65 & \textbf{51.92} \\ 
\bottomrule
\end{tabular}
\end{table}

\subsection{3D Semantic Evaluation}

Fig. \ref{fig:semantic} and Tab. \ref{table:semantic} present the qualitative and quantitative results for 3D zero-shot semantic segmentation. Both 3D-OVS and LERF employ a single MLP to regress semantic features across the entire scene, resulting in blurred instance boundaries and relatively noisy segmentation outcomes. ConceptGraph incrementally merges objects based on geometric and semantic similarities between point cloud segments. However, this approach tends to cause under-segmentation issues, such as merging the mat and the bed in the room\_1. In contrast, our method utilizes a Code Book to store semantic features and constructs a separate NeRF model for each instance, which better preserves instance-level details, maintains clear boundaries between instances. Additionally, we include OpenObj as the single-agent baseline for OpenMulti, with a more detailed comparison provided in next section.


\vspace{-8pt}
\subsection{Multi vs Single}\label{multivssingle}

\begin{table}[t]
  \centering
  \caption{Comparison of "OpenSingle" and OpenMulti}
  \label{tab:multivssingle}
  \setlength{\tabcolsep}{1.7mm}
  \begin{tabular}{lcccccc}
    \toprule
     & \multicolumn{2}{c}{Geometric} & \multicolumn{2}{c}{Semantic} & \multicolumn{1}{c}{Time} \\
    \cmidrule(lr){2-3} \cmidrule(lr){4-5} \cmidrule(lr){6-6}
     & Acc. $\downarrow$ & Comp. $\downarrow$ & F-mIoU $\uparrow$ & F-mAcc $\uparrow$ & Time[min] $\downarrow$ \\
    \midrule
    OpenObj & 3.14 & 4.42 & 44.10 & 49.65 & 12.43 \\
    Ours    & \textbf{2.47} & \textbf{3.52} & \textbf{46.18} & \textbf{51.92} & \textbf{4.76} \\
    \bottomrule
  \end{tabular}
\end{table}

Tab. \ref{tab:multivssingle} compares the mapping accuracy and runtime between the "OpenSingle" (OpenObj) and our method. Relying on a multi-agent distributed mapping paradigm, our method achieves significantly shorter runtime.

In terms of accuracy, OpenMulti also outperforms "OpenSingle", mainly due to the semantic correction enabled by the cross-agent instance alignment module. Unlike "OpenSingle", where early semantic errors can propagate and degrade subsequent predictions, the multi-agent framework corrects biased results by leveraging confidence-based cross-agent collaboration, thereby improving overall semantic consistency.

\vspace{-8pt}
\subsection{Ablation Study}

\begin{figure}[t]
    \centering
    \includegraphics[width=0.48\textwidth]{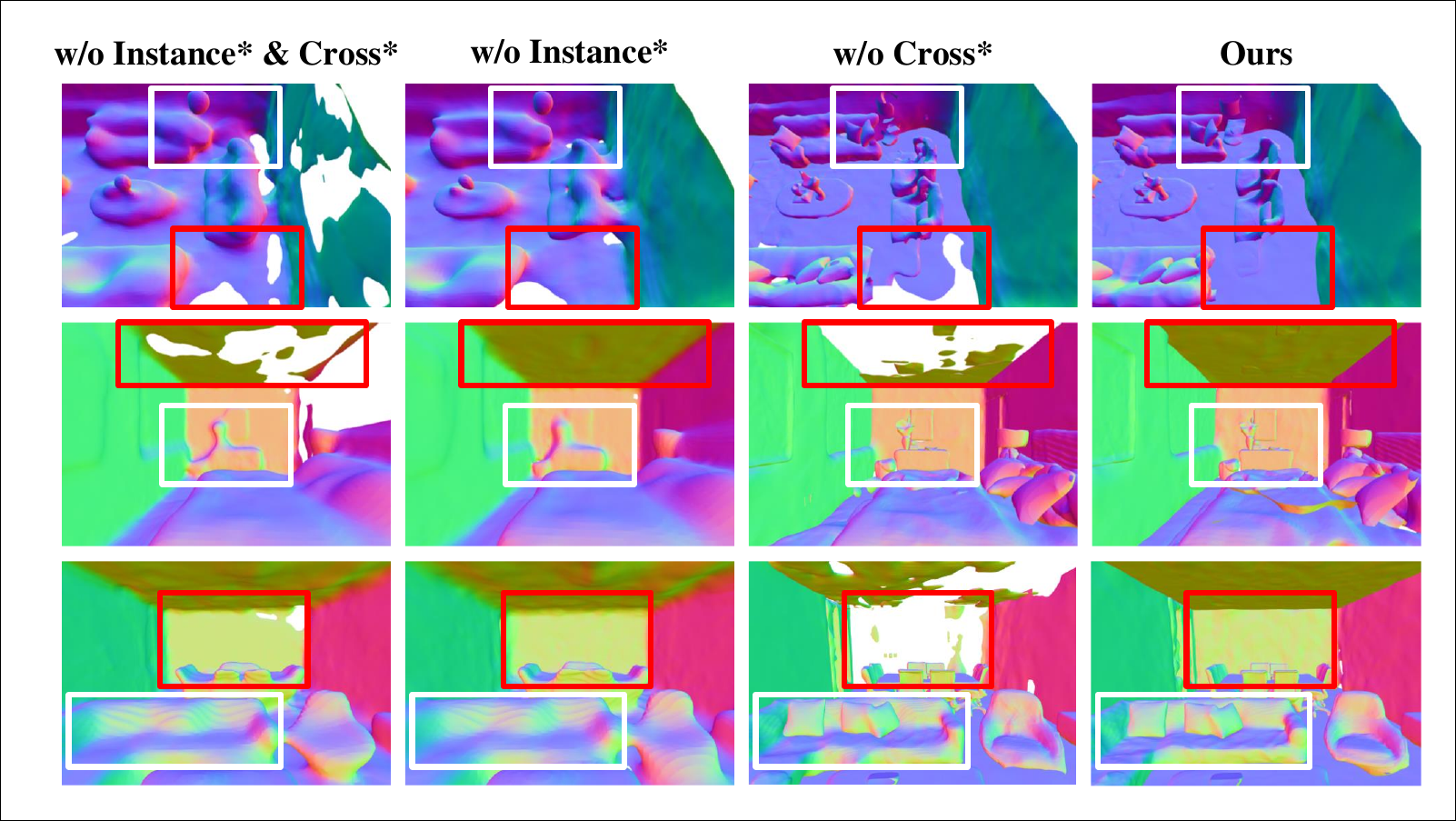}
    \caption{Geometric reconstruction comparison in the ablation study. The red box highlights the influence of Cross Rendering Supervision on scene reconstruction, whereas the white box emphasizes the impact of instance-level mapping.
}
    \label{fig:ablation}
    \vspace{-10pt}
\end{figure}

Tab. \ref{table:ablation}. presents the results of the ablation study. 
W/o Instance* eliminates the instance-level mapping mode, employing the same optimization method to construct a global map. Consequently, all instances are merged, leading to poor detail reconstruction, as shown in the white box in Fig. \ref{fig:ablation}.
W/o Cross* removes the Cross Rendering Supervision, resulting in a significant decline in mapping accuracy, as illustrated in the red box in Fig. \ref{fig:ablation}. 

Additionally, we quantitatively evaluated the overhead introduced by the proposed method. Compared to w/o Cross*, it results in significantly lower communication overhead and only a slight increase in optimization time, while maintaining competitive mapping accuracy. This demonstrates that the module achieves an efficient trade-off between communication overhead, optimization efficiency, and mapping accuracy.

\begin{table}[!t]
\centering
\scriptsize
\caption{Results of Ablation Study}
\label{table:ablation}
\renewcommand\arraystretch{1.1}
\setlength{\tabcolsep}{2.5mm}
\begin{tabular}{cc|cccc}
\toprule
 \textbf{Instance*} & \textbf{Cross*} & \textbf{Acc.$\downarrow$} & \textbf{Comp.$\downarrow$} & \textbf{Time[s]$\downarrow$} & \textbf{Memory[MB]$\downarrow$} \\ \midrule
 $\times$ & $\times$ & 04.08 & 11.26 & \textbf{01.18} & \textbf{02.01}\\
 $\times$ & $\checkmark$ & 03.53 & 08.84 & 02.85 & \textbf{02.01}\\ 
 $\checkmark$ & $\times$ & 02.75 & 03.93 & 05.43 & 02.13\\ 
 $\checkmark$ & $\checkmark$ & \textbf{02.47} & \textbf{03.52} & 05.71 & 02.14\\ \bottomrule
\end{tabular}
\vspace{-5pt}
\end{table}

\subsection{Communication Failure Test}

\par

\begin{figure}[t]
    \centering
    \includegraphics[width=0.48\textwidth]{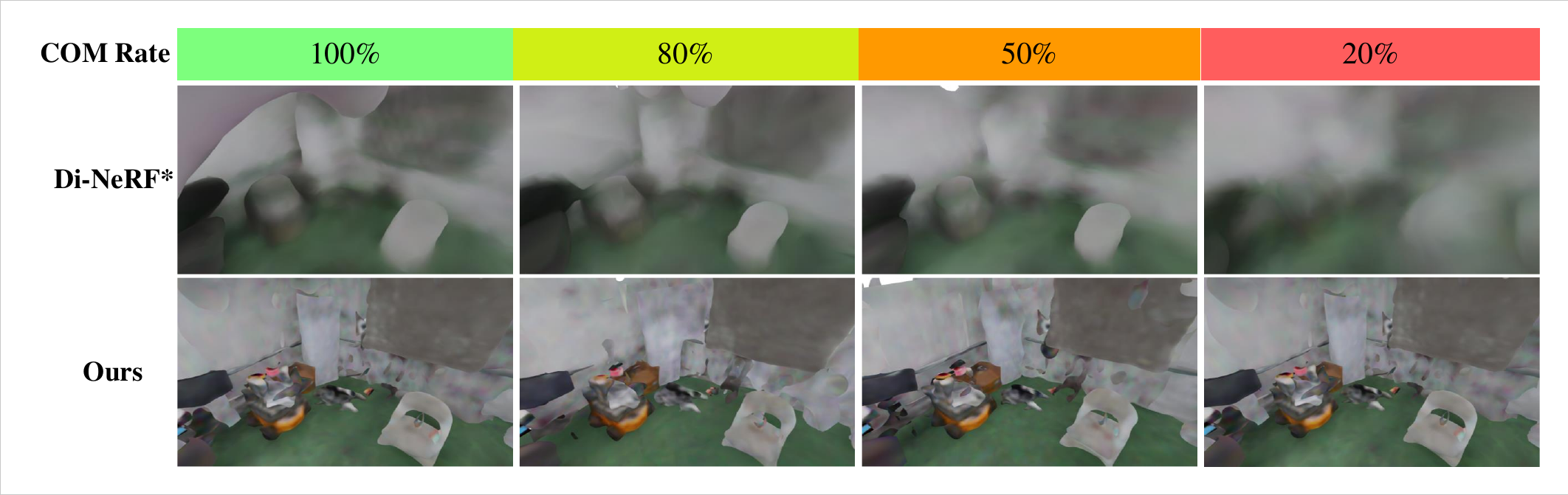}
    \caption{Mapping results under varying communication success rates on real-world data. We collected raw data from four agents in our laboratory environment and simulated communication failures for evaluation.}
    \label{fig:communication_failure}
    \vspace{-10pt}
\end{figure}

We validate the algorithm's robustness through communication failure simulations in the room\_0 scenario. As shown in Tab. \ref{table:failure}, OpenMulti maintains a high mapping completeness rate with only a minor drop (-15.23\%) under degraded communication, outperforming Di-NeRF*'s significant decline (-39.16\%). Relevant qualitative visualization results can be accessed on our website. Fig. \ref{fig:communication_failure} illustrates the mapping performance under varying communication success rates using real-world data. As shown, OpenMulti exhibits greater robustness to communication failures than Di-NeRF*. This is attributed to the instance-level paradigm transmitting complete instance information via single successful communication, whereas global approaches like Di-NeRF* require continuous communication to integrate fragmented scene data, leading to biased parameter learning.

\subsection{Timings and Memory}
Communication cost and optimization time are important metrics for multi-agent collaborative mapping. We selected room\_0 as an example to measure these two parameters for various distributed implicit mapping methods, as shown in Tab. \ref{table:communication}. First, compared with explicit mapping paradigms, their data transmission volume increases exponentially with the expansion of exploration range, resulting in significant communication overhead. When compared to other instance-level mapping methods of the same type, our algorithm demonstrates advantages in optimization time because it eliminates the need for complex backpropagation mechanisms. Relative to global-level mapping methods, the increased number of instances leads to more network parameters in our approach, causing some growth in both data transmission volume and optimization time. However, this brings substantial improvements in mapping accuracy. 



\subsection{Instance-level retrieval}

\begin{figure}[t]
    \centering
    \includegraphics[width=0.48\textwidth]{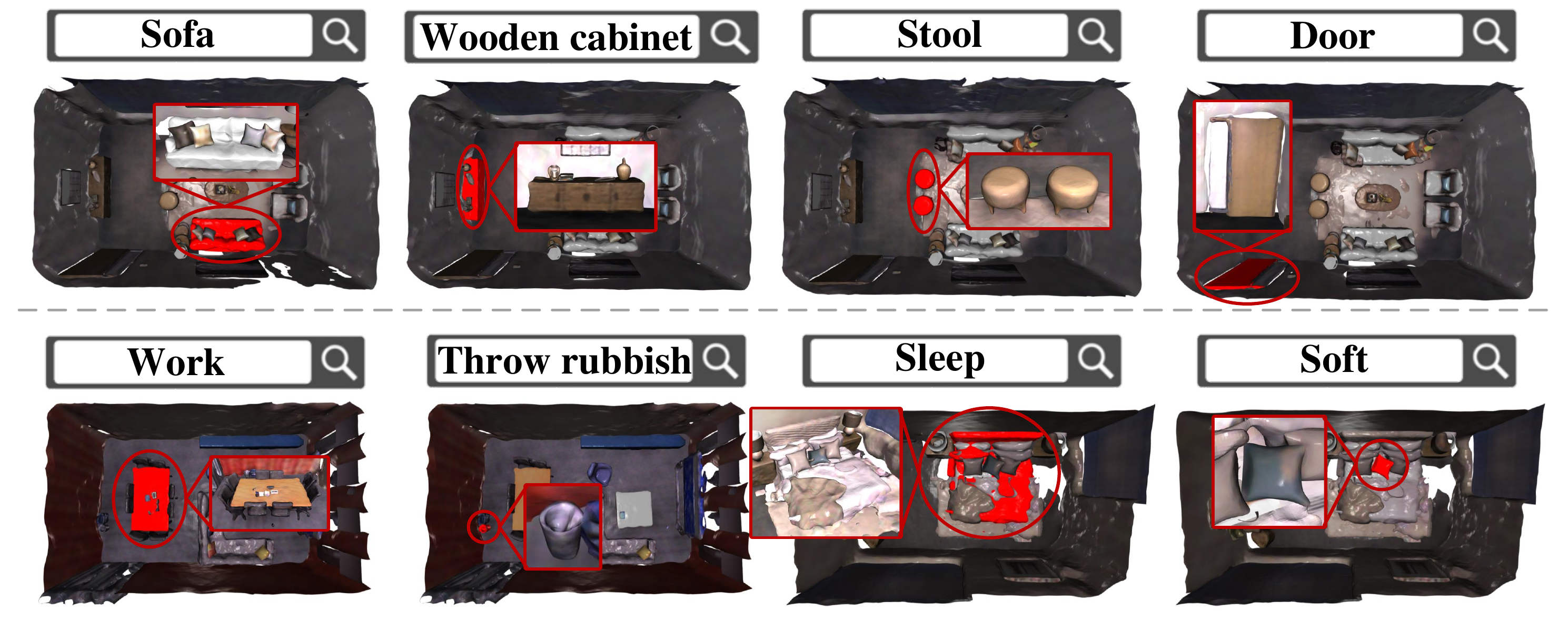}
    \caption{The results of instance retrieval. We conducted some retrieval in various scenes, which yielded significant results.}
    \label{fig:query}
    \vspace{-12pt}
\end{figure}

Fig. \ref{fig:query} presents some results of instance-level retrieval. 
Specifically, we extract CLIP and Caption features from the query text, compute their cosine similarity with semantic features, and identify the queried instance by taking the maximum of their weighted similarity sum. This allows our method to retrieve instances according to their names or functionalities.

\begin{table}[!t]
\centering
\scriptsize
\caption{Completion Ratio under Communication Failure}
\label{table:failure}
\renewcommand\arraystretch{1.1}
\setlength{\tabcolsep}{2mm}
\begin{tabular}{ccccc}
\toprule
 \textbf{COM Rate} & \textbf{100\%} & \textbf{80\%} & \textbf{50\%} & \textbf{20\%} \\ \midrule
 Di-NeRF* & 80.27 & 71.18(-11.30\%) & 65.23(-18.74\%) & 48.84(-39.16\%)\\
 Ours & \textbf{89.36} & \textbf{84.05(-05.94\%)} & \textbf{77.30(-13.50\%)} & \textbf{75.75(-15.23\%)}\\ 
\bottomrule
\end{tabular}
\vspace{-8pt}
\end{table}


\begin{table}[!t]
    \centering
    \caption{Per-frame Transmission and Optimization Costs}
    \label{table:communication}
    \scriptsize
    \setlength{\tabcolsep}{1.6mm}
    \begin{tabular}{ccccc}
        \toprule
        \textbf{} & \textbf{Method} & \textbf{Memory [MB]$\downarrow$} & \textbf{Time [s]$\downarrow$} & \textbf{Acc. [cm]$\downarrow$} \\
        \midrule
        \multirow{1}{*}{Explicit} & Point Clouds & 06.02 $\rightarrow$ 35.08 & -- & -- \\
        \midrule
        \multirow{2}{*}{Geometry-Level} 
        & MACIM & \textbf{01.50 $\rightarrow$ 01.50} & 01.57 & 06.93\\
        & Di-N.* & 02.01 $\rightarrow$ 02.01 & \textbf{01.19} & 03.46 \\
        \midrule
        \multirow{3}{*}{Instance-Level} 
        & Di.I. & 02.13 $\rightarrow$ 03.06 & 06.67 & 03.66\\
        & DS.I. & 02.13 $\rightarrow$ 03.06 & 14.53 & 04.41\\
        & Ours & 02.14 $\rightarrow$ 03.07 & 05.71 & \textbf{02.47}\\
        \bottomrule
    \end{tabular}
    \vspace{-10pt}
\end{table}

\par

\par

\vspace{-8pt}
\section{Conclusion}

This paper presents OpenMulti, an open-vocabulary instance-level multi-agent distributed implicit mapping framework, which achieves consistent mapping across multiple agents. We introduce the Cross-Agent Instance Alignment module, which constructs an Instance Connectivity Graph to unify instance awareness among agents. Additionally, we propose Cross Rendering Supervision to mitigate the blind-zone optimization trap, further enhancing multi-agent implicit mapping accuracy. OpenMulti also supports downstream tasks, such as instance retrieval.

\vspace{-5pt}

\bibliographystyle{Bibliography/IEEEtran}
\bibliography{Bibliography/arxiv}

\vfill

\end{document}